\documentclass[conference]{IEEEtran}
\IEEEoverridecommandlockouts
\usepackage[noadjust]{cite}
\usepackage{amsmath,amssymb,amsfonts}
\usepackage{graphicx}
\usepackage{xcolor}
\def\BibTeX{{\rm B\kern-.05em{\sc i\kern-.025em b}\kern-.08em
    T\kern-.1667em\lower.7ex\hbox{E}\kern-.125emX}}

\usepackage[labelformat=simple]{subcaption}

\usepackage{amsmath}
\usepackage{amsthm}
\usepackage{amssymb}
\usepackage{algorithm}
\usepackage[noend]{algpseudocode}
\usepackage{multirow}
\usepackage{bbm}

\newcommand{\daeuai}{$DAE_{\mathit{uai}}$}    

\newcommand{\citet}[1]{\cite{#1}}
\newcommand{\citep}[1]{\cite{#1}}
    
\begin{document}

\title{Deep Active Learning for Anomaly Detection\\
\thanks{We thank the partial support given by the Project: Models, Algorithms and Systems for the Web (grant FAPEMIG / PRONEX / MASWeb APQ-01400-14), and authors' individual grants and scholarships from CNPq, Fapemig and Kunumi. TP and MM did this work while at Kunumi, TP is now at University of Cambridge and MM is now at DeepMind.}
}

\author{\IEEEauthorblockN{Tiago Pimentel}
\IEEEauthorblockA{
\textit{Kunumi} \\
Belo Horizonte, Brazil\\
tpimentelms@gmail.com}
\and
\IEEEauthorblockN{Marianne Monteiro}
\IEEEauthorblockA{
\textit{Kunumi} \\
Campina Grande, Brazil \\
mariannelinharesm@gmail.com}
\and
\IEEEauthorblockN{Adriano Veloso}
\IEEEauthorblockA{
\textit{CS Dept@UFMG} \\
Belo Horizonte, Brazil \\
adrianov@dcc.ufmg.br}
\and
\IEEEauthorblockN{Nivio Ziviani}
\IEEEauthorblockA{
\textit{CS Dept@UFMG \& Kunumi} \\
Belo Horizonte, Brazil\\
nivio@dcc.ufmg.br}
}

\maketitle

\begin{abstract}
Anomalies are intuitively easy for human experts to understand, but they are hard to define mathematically. Therefore, in order to have performance guarantees in unsupervised anomaly detection, priors need to be assumed on what the anomalies are. By contrast, active learning provides the necessary priors through appropriate expert feedback. Thus, in this work we present an active learning method that can be built upon existing deep learning solutions for unsupervised anomaly detection, so that outliers can be separated from normal data effectively. We introduce a new layer that can be easily attached to any deep learning model designed for unsupervised anomaly detection to transform it into an active method. We report results on both synthetic and real anomaly detection datasets, using multi-layer perceptrons and autoencoder architectures empowered with the proposed active layer, and we discuss their performance on finding clustered and low density anomalies.
\end{abstract}


\section{Introduction} \label{sec:introduction}

Anomaly detection (a.k.a. outlier detection) \cite{hodge2004survey,chandola2009anomaly,aggarwal2015outlier} aims to discover instances that do not conform to the patterns of majority. The key challenge in anomaly detection applications is that sufficient anomalies and correct labels are often prohibitively expensive to acquire.
This problem has been amply studied \citep{liu2017accelerated,zheng2017contextual,zong2018deep}, with solutions inspired by extreme value theory \citep{siffer2017anomaly}, robust statistics \citep{zhou2017anomaly} and graph theory \citep{perozzi2014focused}.

Given that label acquisition is expensive and time consuming, anomaly detection is often applied on unlabeled data which is known as unsupervised anomaly detection. 
It is a specially hard task, since there is no information on what anomalous instances are. Still, there is a rising trend of adopting unsupervised anomaly detection, with most works using models with implicit priors or heuristics to discover anomalies and providing an anomaly score $s(x)$ for each instance in a dataset. Active anomaly detection is a powerful alternative approach to this problem, which has presented good results in recent works \citep{sharma2016active,ref2,veeramachaneni2016ai,das2016incorporating,das2017incorporatingtree}. The basic idea is that experts can give feedback, thus indicating a few anomaly examples. The subset of anomalies provides valuable input in order to learn better representations of what is normal and anomalous.

Since unsupervised anomaly detection requires priors to be assumed on the anomaly distribution, we approach the anomaly detection problem with an active learning method which we call UAI $-$ Unsupervised to Active Inference. UAI is a layer that can be applied on top of any unsupervised anomaly detection deep learning model to transform it into an active model. The UAI layer is a classifier trained on usually few already labeled instances using the strongest assets of deep unsupervised anomaly detection models: the learned latent representations coupled with an anomaly score.

Experimental results show that the UAI layer applied on top of simple deep unsupervised anomaly detection architectures outperforms state-of-the-art anomaly detection methods on several synthetic and real datasets.  We compare the anomaly detection performance of our models against unsupervised, semi-supervised and active competitors under similar budgets. This is done without needing hyper-parameter tuning, as in practical applications, we would not have (initially) any labeled data on which we could tune hyper-parameters, or have too few labeled instances to do it reliably. We also visualize our models' learned latent representations and compare them to the ones from unsupervised models.
\begin{figure*}[t]
  \centering
  \includegraphics[width=0.4\textwidth]{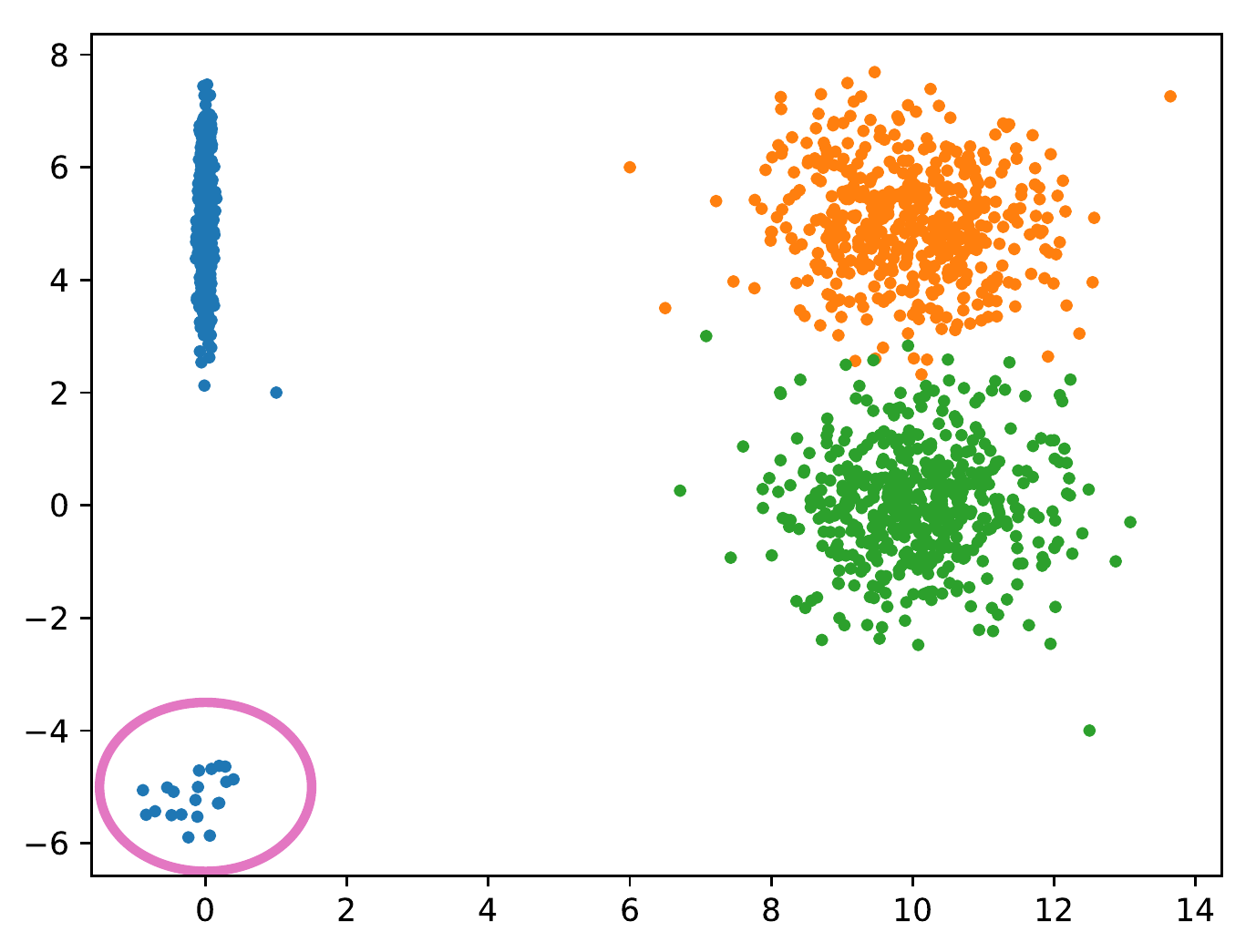}
  \includegraphics[width=0.4\textwidth]{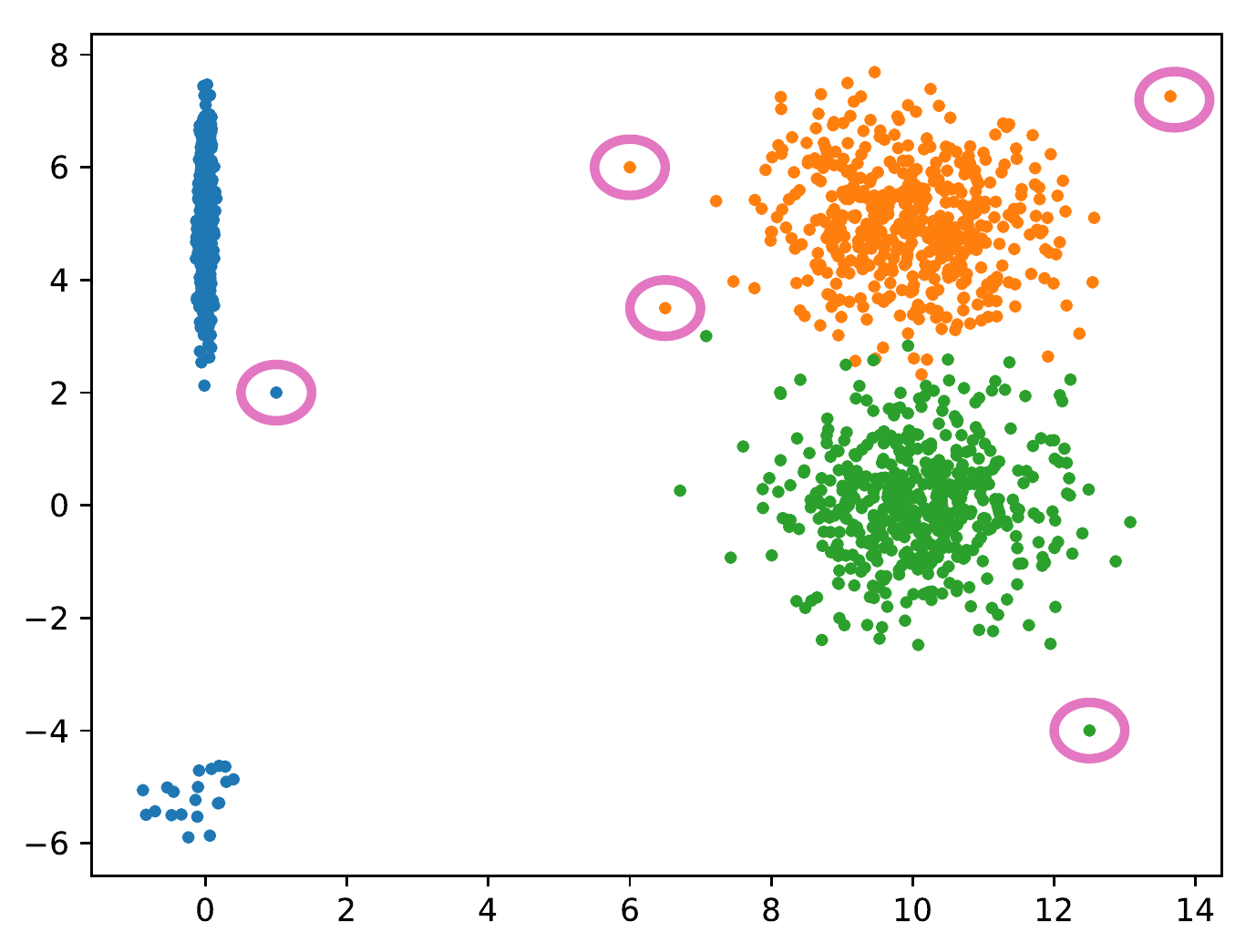}
\caption{Illustrative example of undecidable anomaly distribution. (Left) Anomalies are clustered. (Right) Low density anomalies.}
\label{fig:clusters_anoms}
\end{figure*}

\section{Problem Definition} \label{sec:definition}

In \citet{grubbs1969procedures}, the authors define an outlier as one that appears to deviate markedly from other members of the sample in which it occurs. In~\citet{hawkins1980identification}, the authors state that an outlier is an observation that deviates so much from other observations as to arouse suspicion that it was generated by a different mechanism. Another definition appears in~\citet{chandola2009anomaly}: ``normal'' data points occur in high probability regions of a stochastic model, while anomalies occur in the low probability ones.

Following these definitions, %
we assume there is a probability density function from which ``normal'' data points are generated as
$
p_{\mathit{normal}} \left( x \right) = p \left( x |y=0 \right)
$,
where $x$ is a data point
and $y$ is a label saying if the point is anomalous or not. There is also a different probability density function from which anomalous points are sampled as
$
p_{\mathit{anom}} \left( x \right) = p \left( x |y=1 \right)
$. A full dataset is composed of points sampled from the probability distribution that follows:
\begin{eqnarray}
\nonumber
p_{\mathit{full}} \left( x, y \right) &=&  p \left( y \right) p \left( x | y \right) \\
\nonumber
p_{\mathit{full}} \left( x \right) &=& (1 - \lambda) p_{\mathit{normal}} \left( x \right) + \lambda p_{\mathit{anom}} \left( x \right)
\nonumber
\end{eqnarray}

\noindent where $\lambda$ is an usually small constant representing the probability of a random data point being anomalous.
In~\citet{chandola2009anomaly}, the authors divide anomaly detection learning systems in three different types:

\begin{itemize}
  \item Supervised: A training and a test set are available with curated labels for normal and anomalous points. This case is similar to an unbalanced supervised setting:
  $$ \mathcal{D}_{\mathit{train/test}} = (X, Y)_{\mathit{train/test}} \sim p_{\mathit{full}}(x, y) $$

  \item Semi-Supervised: A training set is available containing only normal points and the task is to identify anomalous points in a test set. This is also called novelty detection:
\begin{eqnarray}
\nonumber
\mathcal{D}_{\mathit{train}} &=& X_{\mathit{train}} \sim p_{\mathit{normal}}(x)  \\
\nonumber
\mathcal{D}_{\mathit{test}} &=& X_{\mathit{test}} \sim p_{\mathit{full}}(x)
\nonumber
\end{eqnarray}

  \item Unsupervised: A dataset containing both normal and anomalous points is available and the task is to identify the anomalous ones:
  $$\mathcal{D} = X \sim p_{\mathit{full}}(x) $$
\end{itemize}

In this work, we focus on \textit{unsupervised anomaly detection}. More specifically, given the full set of points $X \sim p_{\mathit{full}}(x)$, we want to find a subset $X_{\mathit{anom}} \subset X$ containing only anomalous points.
The full distribution $p_{\mathit{full}}$ is a mixture of distributions and
it is a well-known result that general mixture models are unidentifiable \citep{aragam2018identifiability,bordes2006semiparametric}.
Thus, we should not expect to gain information on $p_{\mathit{anom}}$ from knowing $p_{\mathit{full}}$ for any small $\lambda$ without a prior on the anomaly probability distribution, leading us to conclude that unsupervised anomaly detection requires a prior on $p_{\mathit{anom}}$.

Figure~\ref{fig:clusters_anoms} shows a simple example where we illustrate a data distribution composed of three classes of points clustered in four visibly separable clusters. Anomaly detection is an undecidable problem under this setting without further information, since it is impossible to know if the dense cluster is composed of anomalies or the anomalies are the unclustered low density points (or even a combination of both). If we use a low capacity model, the cluster (Figure~\ref{fig:clusters_anoms}  on the Left) would probably present a higher anomaly score. If we use a high capacity model, the low density points (Figure~\ref{fig:clusters_anoms}  on the Right) would be detected as anomalous. Our choice of capacity implicitly imposes a prior on the detected anomalies.

\section{Active Learning Models} \label{sec:model}

The usual strategy to unsupervised anomaly detection is training a parameterized model $p_\theta(x)$ to capture the full data distribution $p_{\mathit{full}}(x)$. Also, since $\lambda$ is by definition a small constant, it is typically assumed that $p_{\mathit{full}}(x) \approx p_{\mathit{normal}}(x)$. Finally, assuming points with low $p_{\mathit{full}}(x)$ values are anomalous \citep{zhou2017anomaly}, an anomaly score is usually defined to be $s(x) \propto \frac{1}{p_\theta(x)}\cdot$ There are three main issues with this strategy:
\begin{itemize}
\item if anomalies are more common than expected, $p_{\mathit{full}}$ might be a poor approximation of $p_{normal}$;
\item if anomalies are tightly clustered in some way, high capacity models may learn to identify that cluster as a high probability region;
\item if anomalies are as rare as expected, and since we only have access to $p_{\mathit{full}}$, we have no information about $p_{anom}$ without further assumptions on its probability distribution.
\end{itemize}

\vspace{0.1in}
\noindent\textbf{Arguments for Active Learning:} The aforementioned issues together argue in favor of an active learning strategy for anomaly detection, including auditor experts in the system's training loop. Thus, anticipating feedback and benefiting from it to find anomalies. Further, having an extremely unbalanced dataset ($\lambda \approx 0$) is another justification for adopting the active learning setting~\cite{ref1,ref4}, which has the potential of requiring exponentially less labeled data than in supervised settings \citep{settles2012active,ref5,ref6}.

\subsection{The UAI Layer}

Unsupervised anomaly detection, by itself, usually presents small accuracy in most practical scenarios, hence it is commonly used to rank instances which are later evaluated by human experts. 

We consider here, then, the task in which we are given a dataset $\mathcal{D} = \{x| x \sim p_{\mathit{full}}(x)\}$, from which possibly anomalous data points are ranked and then sent to be audited by human experts, until a budget $b$ is consumed.\footnote{In these settings, it is not uncommon for large companies to be willing to label considerable sets of data ($b>1000$ instances) in internal auditing processes.} If instead of ranking and selecting all instances once, we iterate with experts in small batches (of $k$ instances each), we can increase the number of anomalies found in these $b$ labeled instances. 

In our active learning setting, then, we iterate with experts. At each step, the $k \ll b$ data points most probably anomalous are sent to be audited and a new training regime takes place once the expert feedback returns.\footnote{We should make it explicit here that this labeling process takes us out of the unsupervised anomaly detection setting. This is also not semi-supervised, though, since we start with no labels. This task is usually denoted as active anomaly detection \citep{das2016incorporating,das2017incorporatingtree}.}
This strategy of selecting the top $k$ elements at each step is called most-likely positive. It is a common approach for selecting informative instances in highly imbalanced datasets \citep{bilgic2012active,sharma2016active}, and follows recent work in active anomaly detection \citep{veeramachaneni2016ai,das2016incorporating,das2017incorporatingtree}.

With this in mind, we develop the UAI $-$ Unsupervised to Active Inference layer. This layer can be incorporated on top of any unsupervised deep learning anomaly detection model which provides an anomaly score for ranking anomalies (e.g., a denoising auto-encoder). It takes as input both a latent representation layer ($l(x)$), created by the model, and its output anomaly score ($s(x)$), and passes it through a classifier to find an item's anomaly score, which is formally defined as:
\begin{equation}
  \hat{p}(y | x) \propto s_{\mathit{uai}}(x) = \mathit{classifier}([l(x);s(x)])
\nonumber
\end{equation}

\noindent where $\hat{p}(y | x)$ is our empirical estimate of the probability of point $x$ being anomalous. This is motivated by recent work stating learned representations have a simpler statistical structure \citep{bengio2013better}, which makes the task of modeling this manifold and detecting unnatural points much simpler \citep{lamb2018fortified}. In this work, we model the UAI layer using a simple logistic regression as our classifier, but any other classifier could be used as well. The classifier is thus given as:
\begin{equation}
\label{eq2}
  \hat{p}(y | x) \propto s_{\mathit{uai}}(x) = \sigma(W_{act}[l(x);s(x)] + b_{act})
\end{equation}
\noindent where $W_{act} \in \mathbb{R}^{1, d+1}$ is a linear transformation, $b_{act} \in \mathbb{R}$ is a bias term and $\sigma(\cdot)$ is the sigmoid function. We learn the values of $W$ and $b$ using back-propagation with a cross entropy loss function, where the targets are the few already actively labeled instances. We allow the gradients to flow through $l$, but not through $s$, since $s$ might be non-differentiable. Hereafter we refer to networks that have an UAI layer as UAInets.

\subsection{Architectures}

Our models were obtained by incorporating the proposed UAI layer into two different anomaly detection architectures.

\vspace{0.05in}
\noindent\textbf{Denoising Autoencoder:} The first model consists of a Denoising Auto-Encoder (DAE). Specifically, an encoder transforms the input into a latent space, and a decoder reconstructs the input using this latent representation. The loss function minimizes the reconstruction error $L_2$ norm. A denoising auto-encoder may be formally defined as follows:
\begin{eqnarray}
\nonumber
  l & = & f_{enc}(x + \epsilon),~ ~ ~ ~\epsilon \sim \mathcal{N}(0, \varphi) \\
\nonumber
  \hat{x} & = & f_{dec}(l) \\
\nonumber
  \mathcal{L}_{\mathit{DAE}} & = & ||x - \hat{x}||_2^2
\nonumber
\end{eqnarray}
\noindent where both $f_{enc}$ and $f_{dec}$ are usually feed forward networks with the same number of layers, $l \in \mathbb{R}^d$ is a $d$-dimensional latent representation and $\epsilon$ is a zero mean noise, sampled from a Gaussian distribution with a $\varphi$ standard deviation. When used in anomaly detection, the reconstruction error is used as an approximation for the anomaly score, as follows:
\begin{equation}
  s_{\mathit{DAE}}(x) = ||x - \hat{x}||_2^2
\nonumber
\end{equation}

\begin{figure}[t]
	\center
        \footnotesize
	\includegraphics[width=\columnwidth]{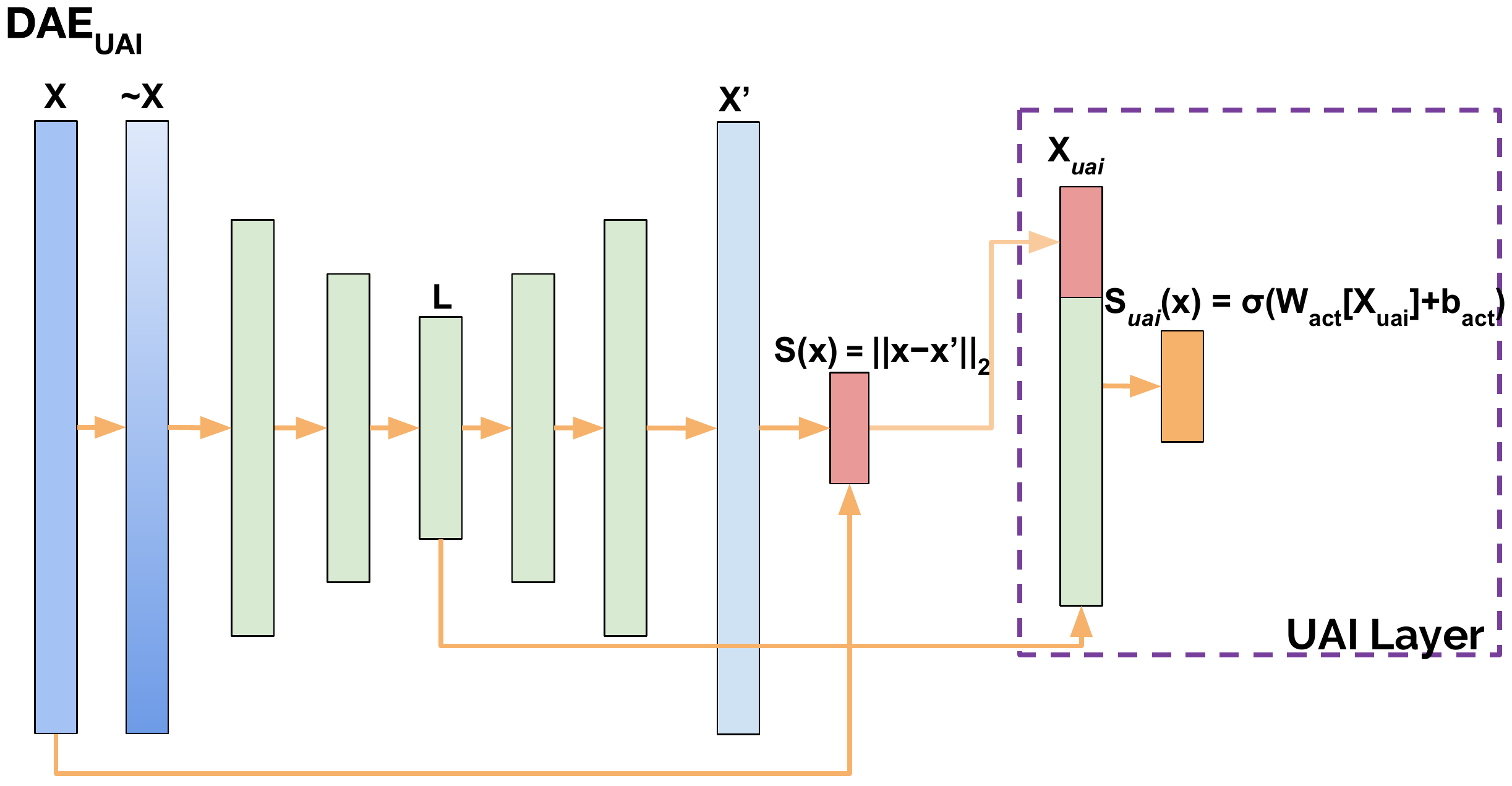}
	\caption{A denoising auto-encoder with an UAI layer.}\label{fig:daeuai_architecture}
\end{figure}

Figure \ref{fig:daeuai_architecture} presents a $\mathit{DAE}_{uai}$ network which assembles the UAI layer on top of the autoencoder. This is formally defined as follows:
\begin{eqnarray}
\nonumber
  l_{\mathit{DAE}} & = & l = f_{enc}(x + \epsilon)\\
\nonumber
  s_{\mathit{DAE}_{uai}}(x) & = & uai([l_{\mathit{DAE}};s_{\mathit{DAE}}])
\nonumber
\end{eqnarray}
\noindent where $uai(\cdot)$ is simply the logistic classifier in Equation~\ref{eq2}. 

\vspace{0.05in}
\noindent\textbf{Multi-Layer Perceptron:} Another unsupervised anomaly detection approach is training a multi-layer perceptron ($Class$) to predict an instance's class label ($x_y$) from its other features ($x_x$).\footnote{Note that even though class labels are available ($x_y$), we initially still have no anomaly label ($y$).} The cross-entropy of a data point serves as an estimate of its anomaly score, as follows:
\begin{eqnarray}
\nonumber
  \widehat{x}_y & = & f_{\mathit{Class}}(x_x) \\
\nonumber
  \mathcal{L}_{\mathit{Class}} & = & H(x_y, \widehat{x}_y) = \mathit{cross\_entropy}\,(x_y, \widehat{x}_y) \\
\nonumber
  s_{\mathit{Class}}(x) & = & H(x_y, \widehat{x}_y)
\nonumber
\end{eqnarray}
\noindent where $f_{\mathit{Class}}(\cdot)$ is a $p-$layer neural network. 

\begin{figure}[t]
	\center
	\footnotesize
	\includegraphics[width=\columnwidth]{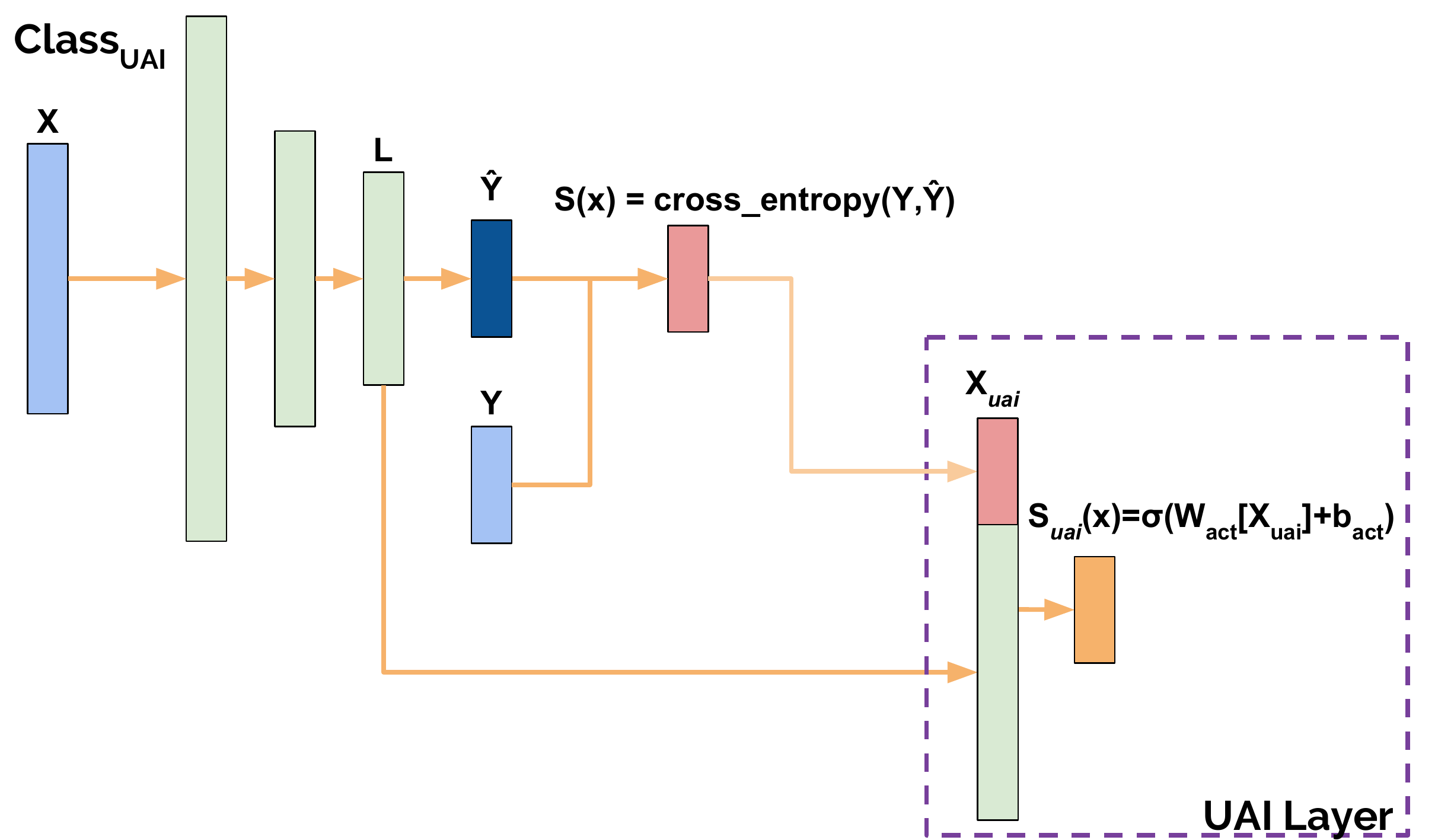}
        \caption{A multi-layer perceptron with an UAI layer.}\label{fig:classuai_architecture}
\end{figure}

Figure~\ref{fig:classuai_architecture} presents $\mathit{Class_{uai}}$, which uses this classifier's last hidden layer ($h_{p-1}$) as a latent representation:
\begin{eqnarray}
\nonumber
  l_{\mathit{Class}} & = & h_{p-1} \\
\nonumber
  s_{\mathit{Class}_{uai}}(x) & = & uai([l_{\mathit{Class}};s_{\mathit{Class}}])
\nonumber
\end{eqnarray}

The full loss function used to train UAInets is:
\begin{eqnarray}
\nonumber
  \mathcal{L}_{\mathit{uai}} & = & \mathbbm{1}~H(y, s_{\mathit{uai}}(x)) \\
\nonumber
  \mathcal{L}_{\mathit{full}} & = & \mathcal{L}_{\mathit{uai}} + \mathcal{L}_{\mathit{base}}
\nonumber
\end{eqnarray}

\noindent where $\mathbbm{1}$ has value $1$ for already actively labeled instances, and 0 for unlabeled ones. $\mathcal{L}_{\mathit{base}}$ refers to the base network's loss function, which will be either $\mathcal{L}_{\mathit{DAE}}$ or $\mathcal{L}_{\mathit{Class}}$ here.
\section{Experiments} \label{sec:experiments}

\begin{figure*}
  \centering
    \begin{subfigure}[t]{0.28\textwidth}
        \centering
        \includegraphics[width=\textwidth]{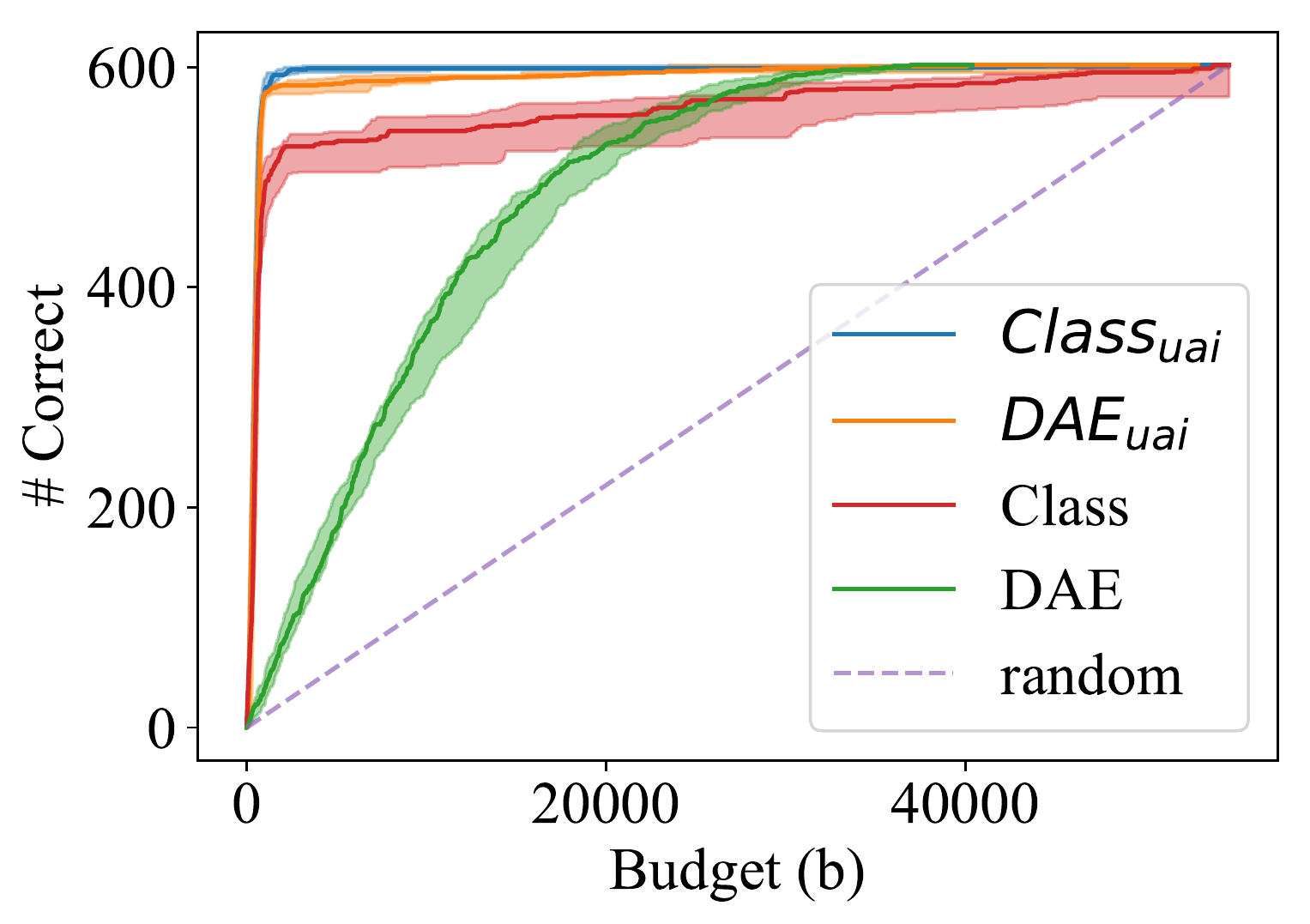}
        \caption{MNIST\textsubscript{0}} \label{fig:anoms_0_all}
    \end{subfigure}%
    \begin{subfigure}[t]{0.28\textwidth}
        \centering
        \includegraphics[width=\textwidth]{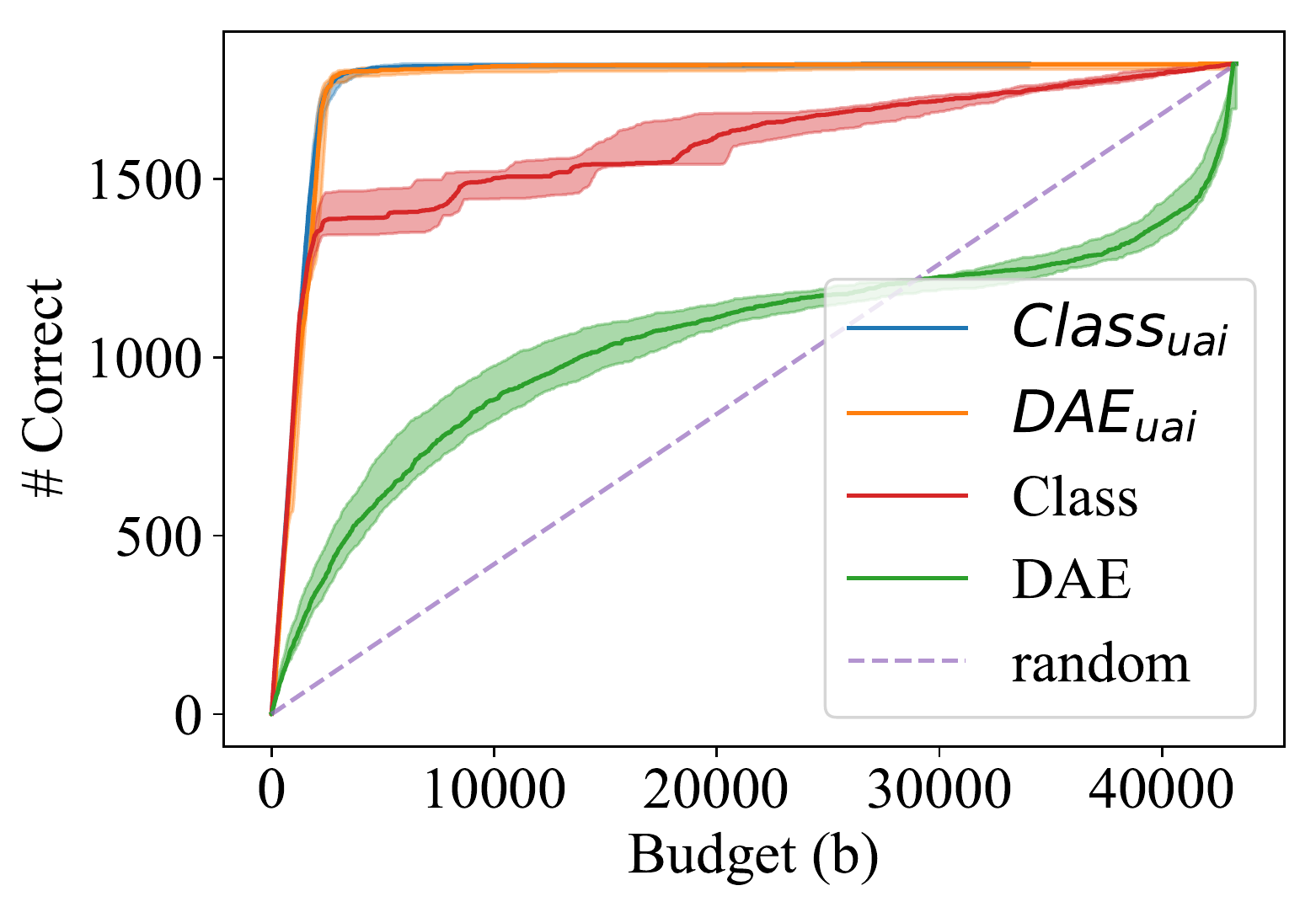}
        \caption{MNIST\textsubscript{0-2}} \label{fig:anoms_012_all}
    \end{subfigure}
    \begin{subfigure}[t]{0.28\textwidth}
        \centering
        \includegraphics[width=\textwidth]{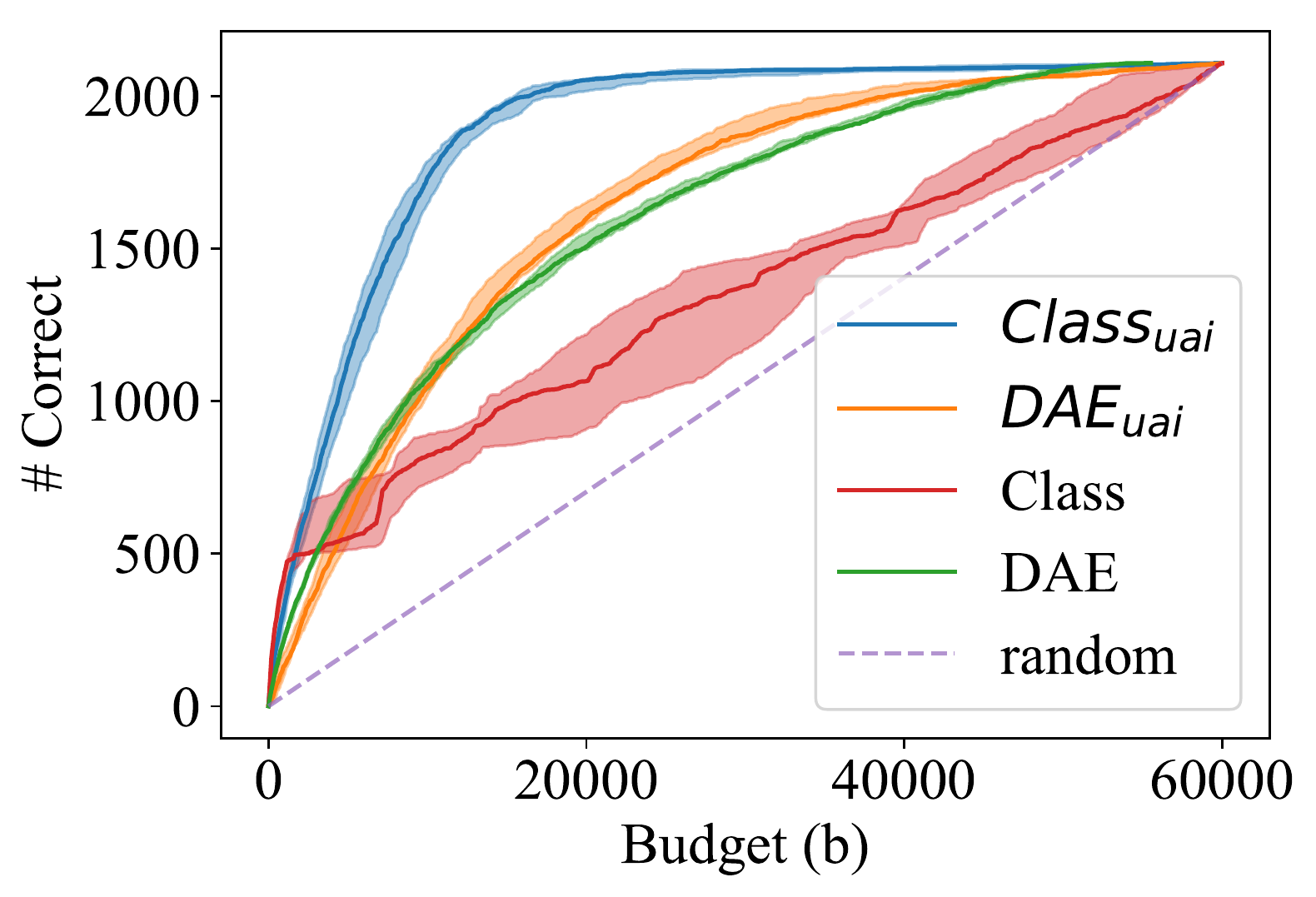}
        \caption{MNIST\textsubscript{hard}} \label{fig:anoms_hard_all}
    \end{subfigure}
  \caption{(Color online) Results for different MNIST experiments. Curves represent the average of five runs with different seeds. Confidence intervals represent max and min results for each budget $b$. $y$-axis represents number of anomalies detected with a specific budget. For MNIST$_0$, with a budget $b=800$ we find almost all $600$ anomalies with both UAInets, but less than $500$ with $\mathit{Class}$ and only $30$ with $\mathit{DAE}$.}
  \label{fig:anoms_all}
\end{figure*}

In this section, we evaluated our active learning models for anomaly detection. We assess the performance of the models by analyzing them on synthetic data created with different properties (Section \ref{sec:synthetic_data}) and on real anomaly data (Section \ref{sec:real_data}).

\subsection{Setup}

For all experiments in this work the DAE's encoder and decoder have independent weights and we used both the $\mathit{DAE}$ and $\mathit{Class}$ models with $3$ hidden layers with sizes $[256, 64, 8]$. This means the latent representations provided to the UAI layers are $l \in \mathbb{R}^8$. 

Algorithm \ref{alg:active_anom} presents our Active Anomaly Detection process. 
We implemented all architectures using TensorFlow \citep{abadi2016tensorflow}. We set the learning rate to $0.01$, batch size to 256 and we used the RMSprop optimizer with the default hyper-parameters.\footnote{Hyper-parameters were hand picked based on a few initial synthetic experiments on MNIST and not tuned in any further way.} We pre-trained the $\mathit{DAE}$ and $\mathit{Class}$ models (by themselves, fully unsupervised and with no active sampling) for $5{,}000$ optimization steps. After that, for the active detection models, we select $k=10$ data points to be labeled (i.e. normal or anomalous) at a time, and further train the full model for $100$ iterations after each labeling call.

\begin{algorithm}
\caption{Active Anomaly Detection}\label{alg:active_anom}
\begin{algorithmic}[1]
\Procedure{ActiveAnomalyDetection} {$\mathcal{D}$, expert, $b$, $k=10$, $steps_{pre}=5000$, $steps_{active}=100$}
\State $\textit{model}\text{.pretrain}(steps_{pre}, \mathcal{D})$
\State $i \gets 0$
\State $\textit{labels} \gets \emptyset$
\While{$i<b$}
  \State $\textit{model}\text{.train}(steps_{active}, \mathcal{D}, \textit{labels})$
  \State $\textit{top\_k} \gets \textit{model}\text{.select\_top}(k, \mathcal{D}, \textit{labels})$
  \State $\textit{labels} \gets \textit{labels} \cup \textit{expert}\text{.audit}(\textit{top\_k})$
  \State $i \gets i + k$
\EndWhile
\EndProcedure
\end{algorithmic}
\end{algorithm}

\subsection{Synthetic Data} \label{sec:synthetic_data}

\begin{figure*}
  \centering
    \begin{subfigure}[t]{0.28\textwidth}
        \centering
        \includegraphics[width=\textwidth]{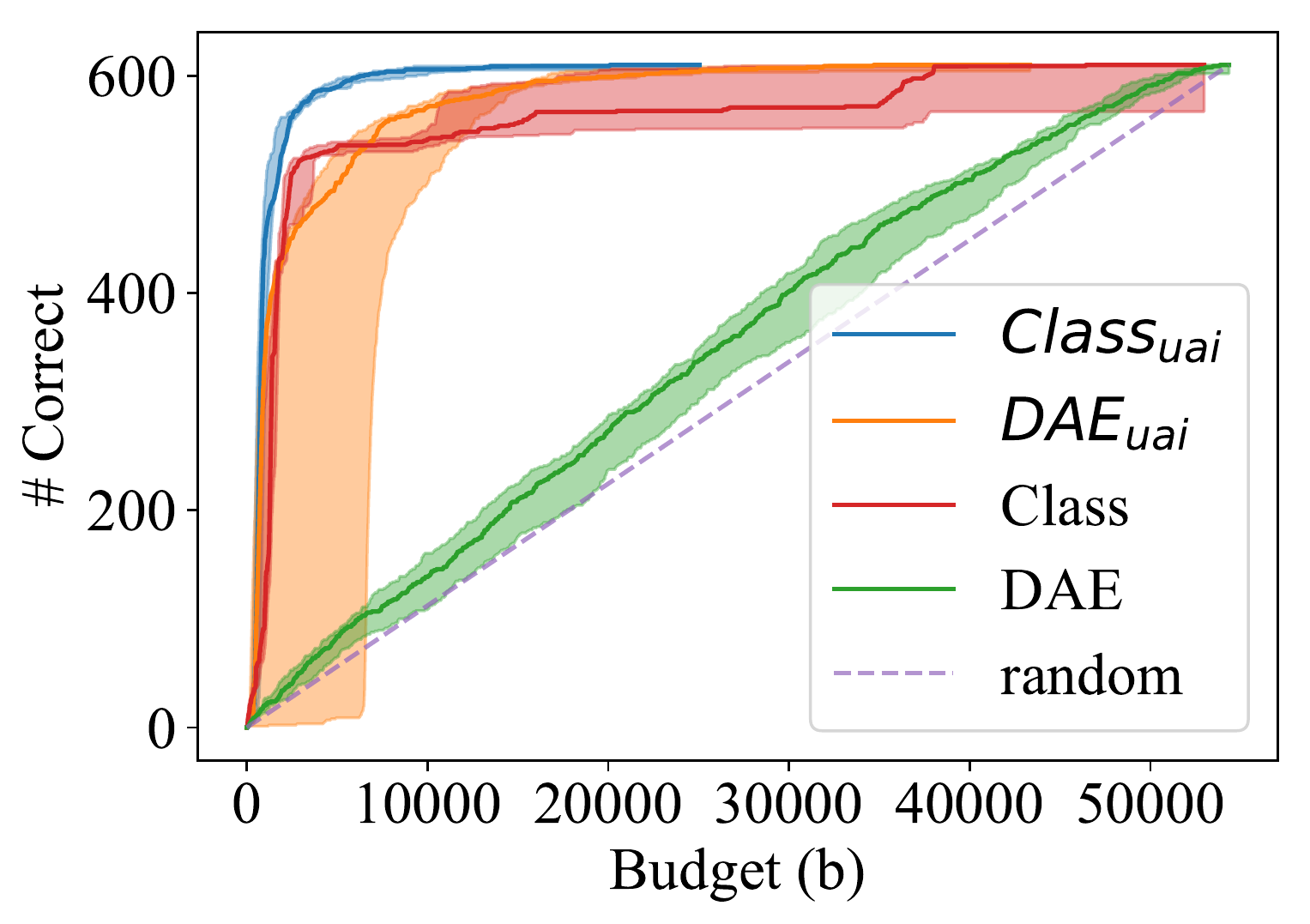}
        \caption{Fashion-MNIST\textsubscript{0}} \label{fig:anoms_0_all_fashion}
    \end{subfigure}%
    \begin{subfigure}[t]{0.28\textwidth}
        \centering
        \includegraphics[width=\textwidth]{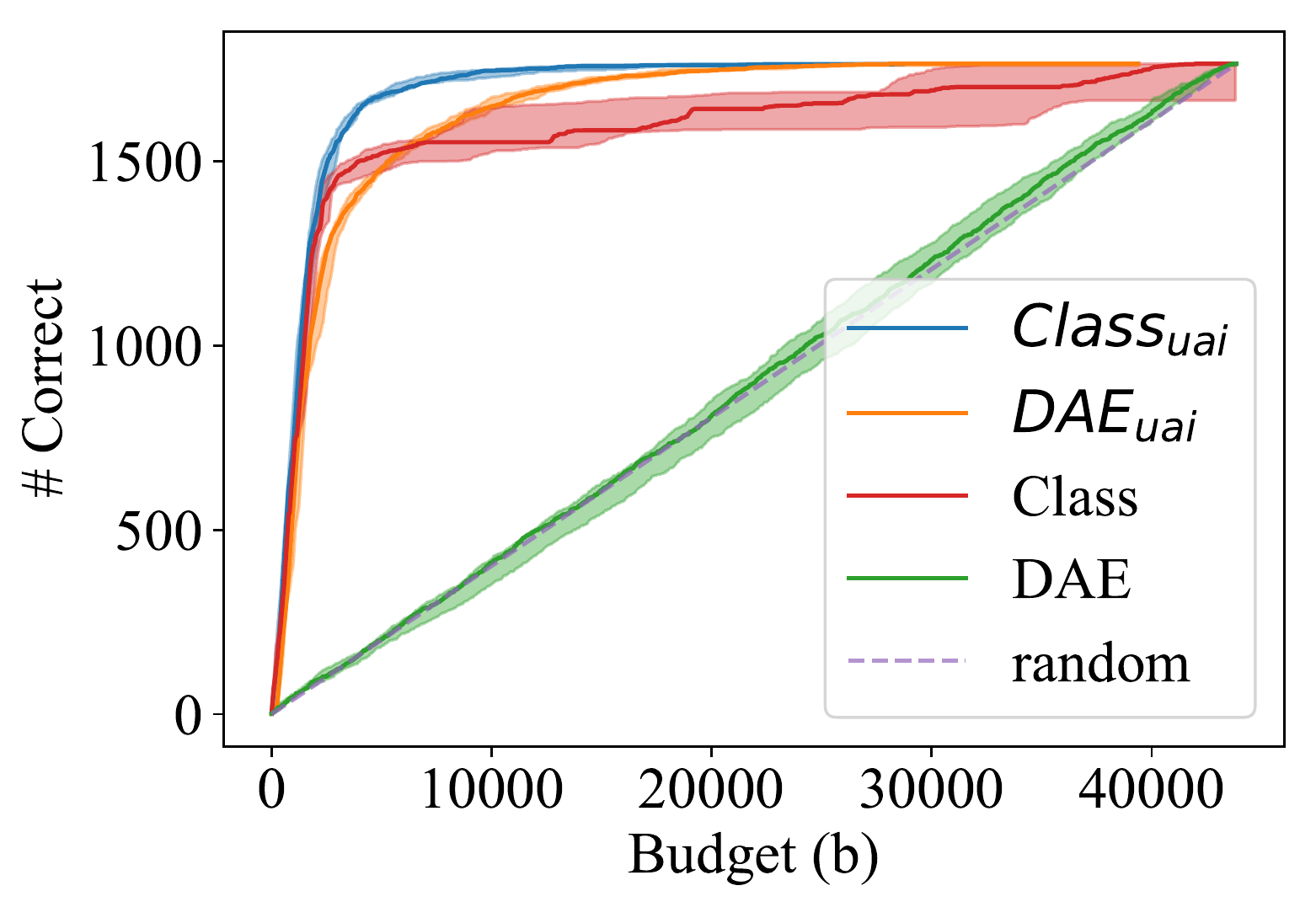}
        \caption{Fashion-MNIST\textsubscript{0-2}} \label{fig:anoms_012_all_fashion}
    \end{subfigure}
    \begin{subfigure}[t]{0.28\textwidth}
        \centering
        \includegraphics[width=\textwidth]{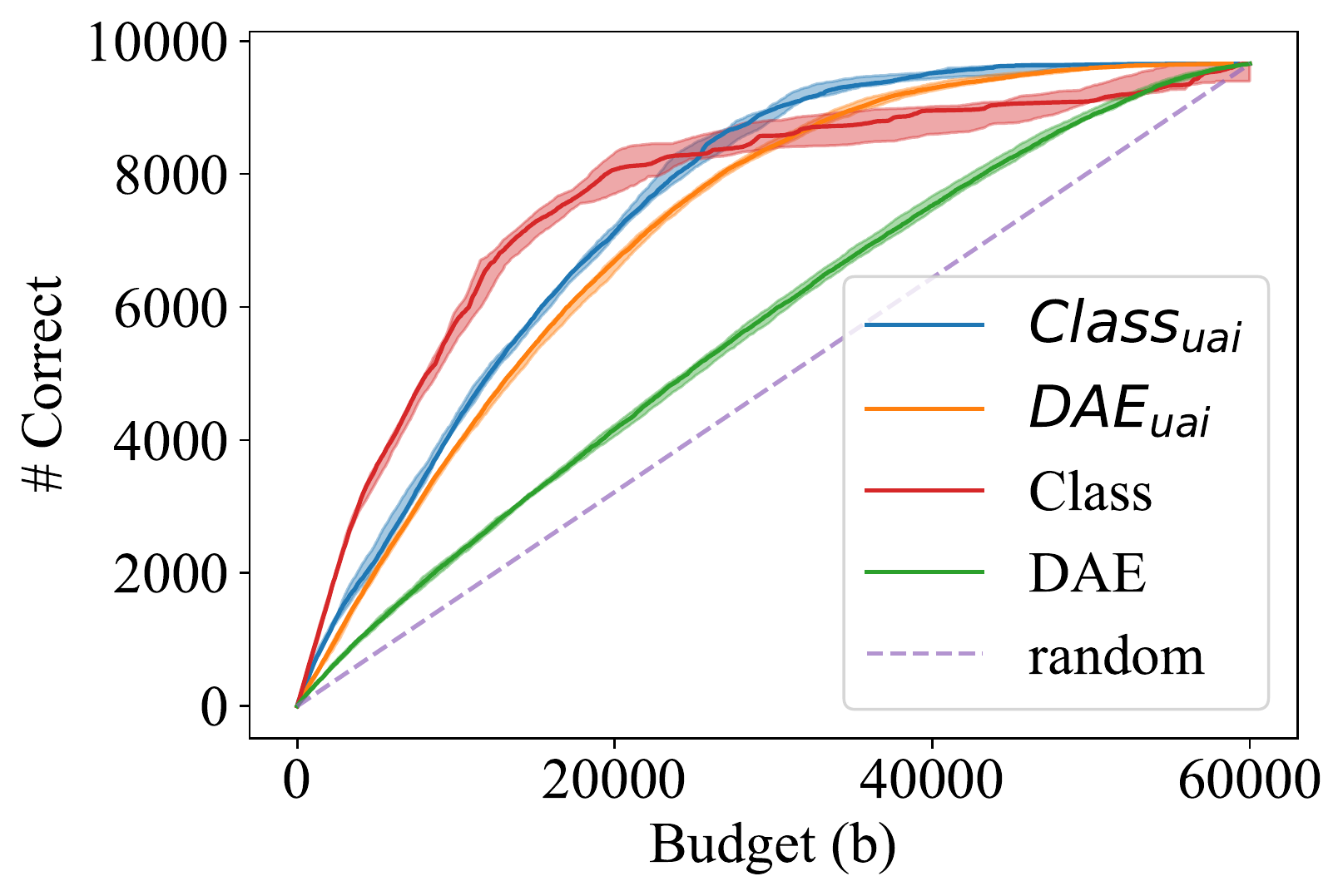}
        \caption{Fashion-MNIST\textsubscript{hard}} \label{fig:anoms_hard_all_fashion}
    \end{subfigure}
    ~ 
    \begin{subfigure}[t]{0.28\textwidth}
        \centering
        \includegraphics[width=\textwidth]{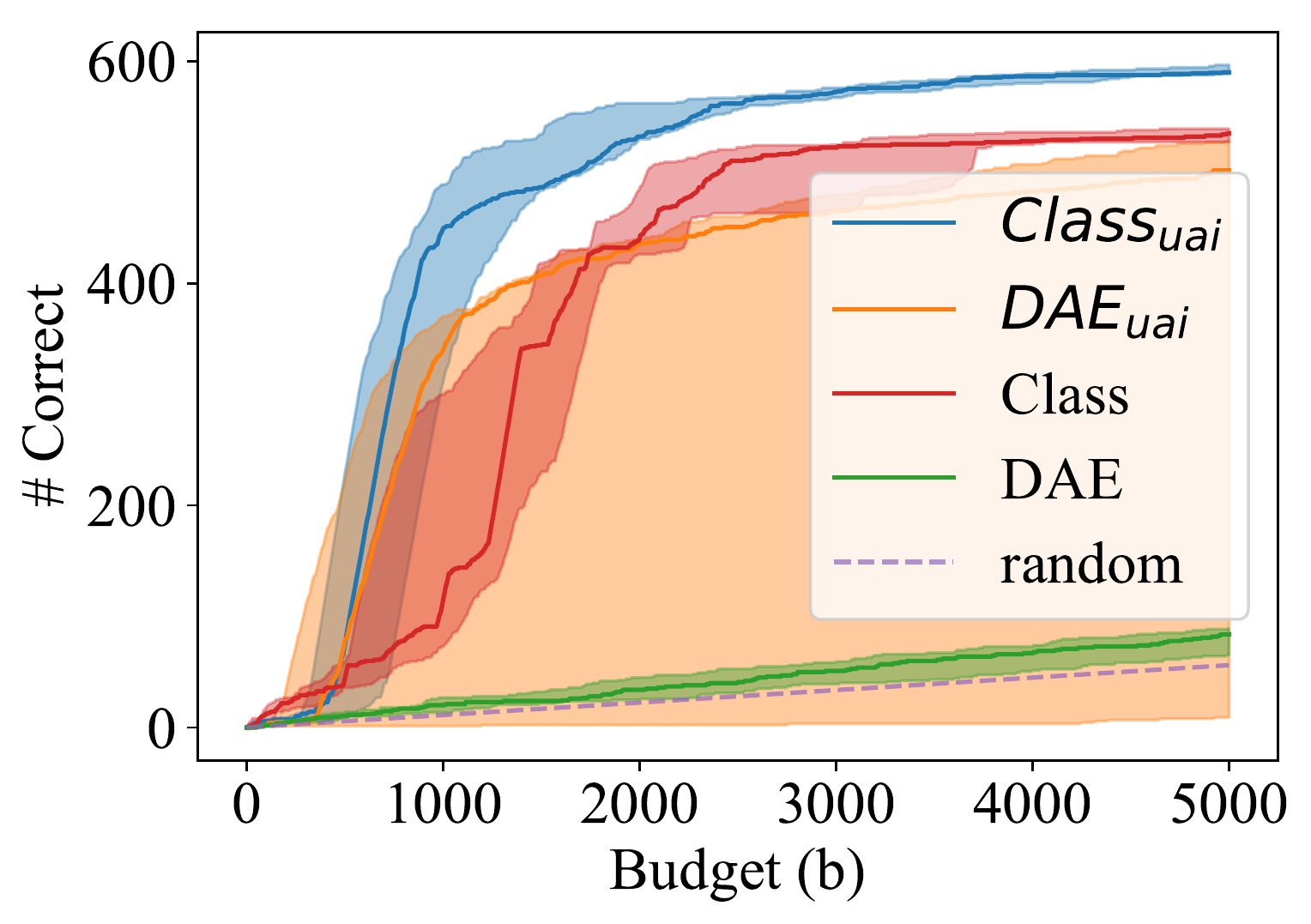}
        \caption{Fashion-MNIST\textsubscript{0} ($b<5000$)} \label{fig:anoms_0_all_zoom_fashion}
    \end{subfigure}
    \begin{subfigure}[t]{0.28\textwidth}
        \centering
        \includegraphics[width=\textwidth]{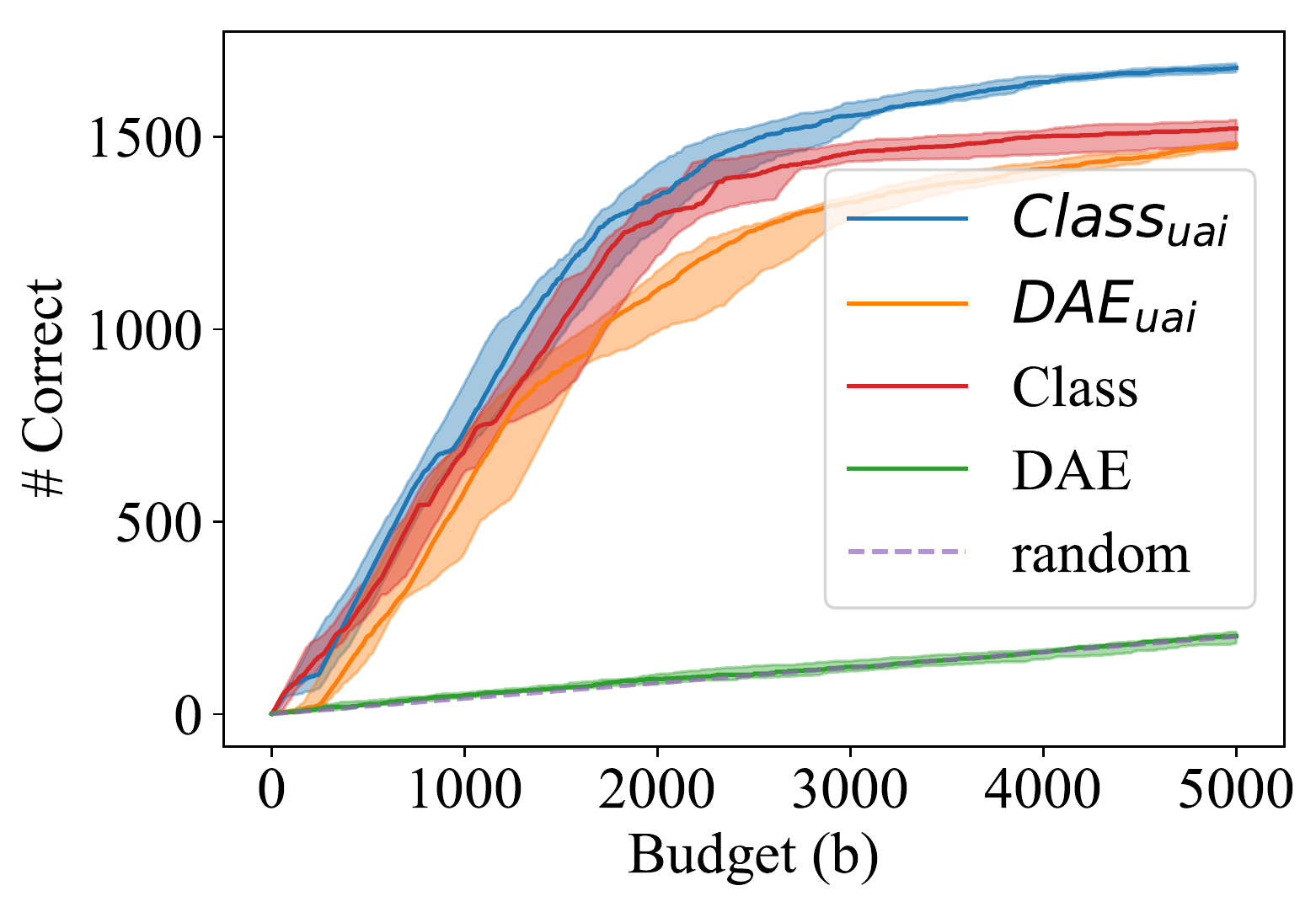}
        \caption{Fashion-MNIST\textsubscript{0-2} ($b<5000$)} \label{fig:anoms_012_all_zoom_fashion}
    \end{subfigure}
    \begin{subfigure}[t]{0.28\textwidth}
        \centering
        \includegraphics[width=\textwidth]{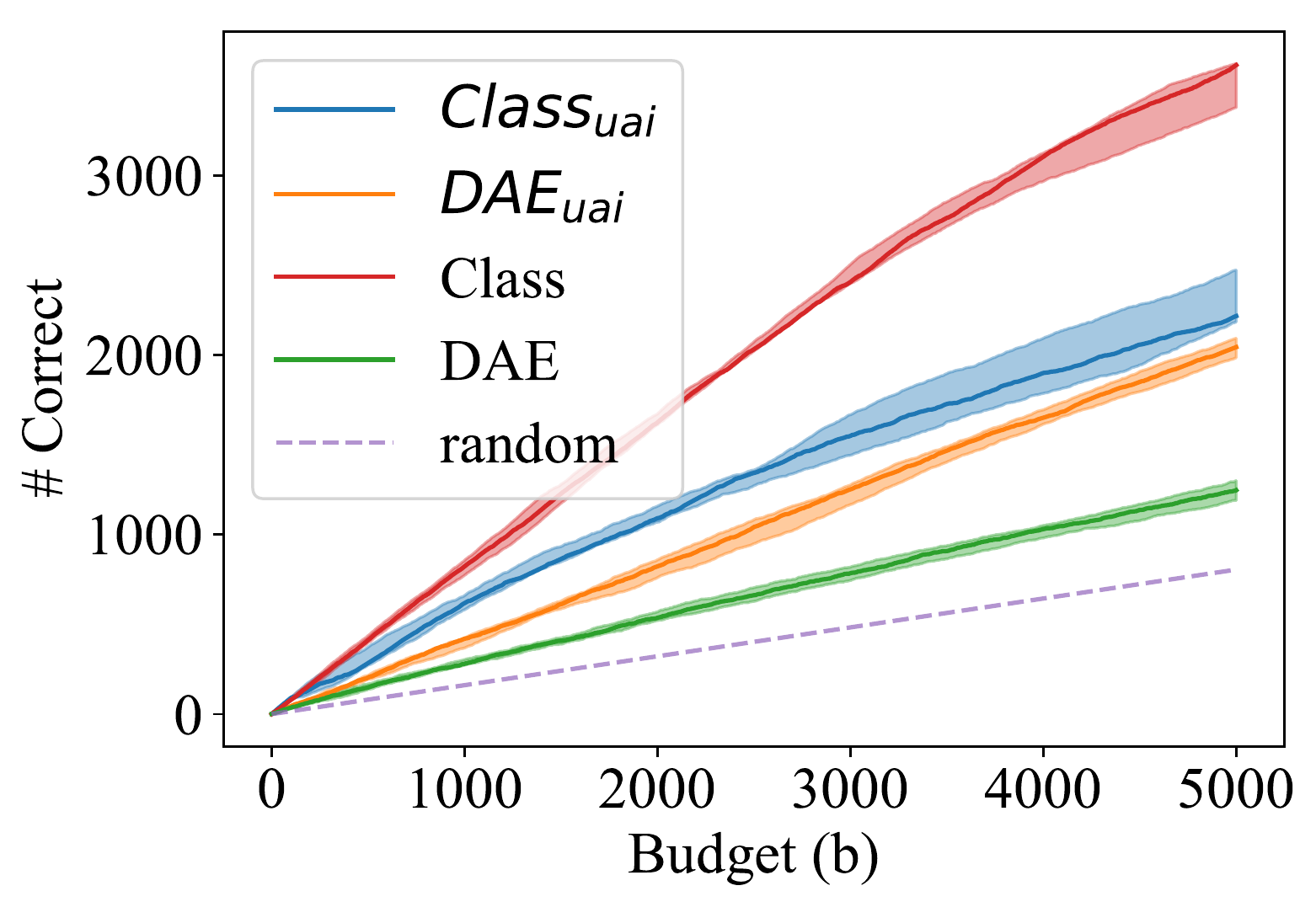}
        \caption{Fashion-MNIST\textsubscript{hard} ($b<5000$)} \label{fig:anoms_hard_all_zoom_fashion}
    \end{subfigure}
  \caption{(Color online) Results for Fashion-MNIST experiments, with different zooms on x-axis. Curves represent the average of five runs with different seeds. Confidence intervals represent max and min results for each budget $b$.}
  \label{fig:anoms_all_zoom_fashion}
\end{figure*}

When designing our experiments, we had the objective of showing that our model can work with different definitions of what is anomalous, while completely unsupervised models will need, by definition, to trade-off accuracy in one setting for accuracy in others.

\vspace{0.05in}
\noindent\textbf{MNIST Datasets:} We used the MNIST dataset\footnote{Using MNIST for the generation of synthetic anomaly detection datasets follows the same strategy as recent works \citep{zhai2016deep,zhou2017anomaly}.} and defined three sets of experiments:

\begin{enumerate}
  \item \textbf{MNIST\textsubscript{0}}: For the first set of experiments, we reduced the presence of the digits with the $0$ label to only $10\%$ of its original number of samples, making it only $1/91 \approx 1.1\%$ of the data points. The $0$s still present in the dataset had their label randomly changed to $y \sim \mathit{Uniform}([1;9])$ and were defined as anomalies.
  \item \textbf{MNIST\textsubscript{0-2}}: Follows the same dataset construction, but we reduce the number of instances of digits with labels $0$, $1$ and $2$, changing their labels to $y \sim \mathit{Uniform}([3;9])$, and again defining them as anomalous. Anomalies composed $3/73 \approx 4.1\%$ of the dataset.
  \item \textbf{MNIST\textsubscript{hard}}: This set of experiments aims to test a different type of anomaly. In order to create the corresponding dataset, we first trained a weak one hidden-layer classifier on MNIST and selected all misclassified data points as anomalous. Anomalies composed $\approx 3.3\%$ of the dataset.
\end{enumerate}

Figure \ref{fig:anoms_all} presents results for these experiments and our main conclusion is that our models are robust to different types of anomalies, which is not the case for the unsupervised models. While $\mathit{Class}$ achieves good results in MNIST\textsubscript{0} and MNIST\textsubscript{0-2} datasets, it does not achieve the same performance in MNIST\textsubscript{hard}, which might indicate it is better at finding clustered anomalies than low density ones. At the same time, $\mathit{DAE}$ achieves good results for MNIST\textsubscript{hard}, but performed poorly on MNIST\textsubscript{0} and MNIST\textsubscript{0-2}, which indicates it is better at finding low density anomalies than clustered ones. Nevertheless, both UAInets are robust in all three experiments.

\vspace{0.05in}
\noindent\textbf{Fashion-MNIST Datasets:} Table \ref{table:mnist-fashion_data_statistics} presents statistics of datasets used to perform experiments on synthetic anomaly detection datasets based on Fashion-MNIST \citep{xiao2017fashion}. To create these datasets we follow the same procedures as with MNIST datasets, thus generating Fashion-MNIST\textsubscript{0}, Fashion-MNIST\textsubscript{0-2}, and Fashion-MNIST\textsubscript{hard}.

\begin{table}[htp]
\center
\caption{Fashion-MNIST Anomaly Datasets.} \label{table:mnist-fashion_data_statistics}
\setlength\tabcolsep{4pt}
\begin{tabular}{lrrrr}
& Dimension & \# classes & \# points & \% anomalies \\
\hline\\
Fashion-MNIST\textsubscript{0} & 784 & 9 & $54{,}610$ & 1.1\% \\
Fashion-MNIST\textsubscript{0-2} & 784 & 7 & $43{,}765$ & 4.0\% \\
Fashion-MNIST\textsubscript{hard} & 784 & 10 & $60{,}000$ & 16.1\% \\
\end{tabular}
\end{table}

Figure~\ref{fig:anoms_all_zoom_fashion} shows results of experiments performed on these datasets following the same procedures as with MNIST datasets. This figure shows similar trends to the ones for MNIST, although the anomalies in these datasets seem harder to identify. In one run of Fashion-MNIST\textsubscript{0}, \daeuai{} needed several active feedback iterations to start learning,\footnote{Figures~\ref{fig:anoms_all} and~\ref{fig:anoms_all_zoom_fashion} confidence curves represent min and max results over five experiments with each model on each dataset. In Figure~\ref{fig:anoms_all_zoom_fashion}(d), in one of the experiments the autoencoders took a long time to find its first anomalous instance. It only found normal instances on its first 5000 picks, so the UAI layer had no signal from which to improve until then.} while for Fashion-MNIST\textsubscript{hard}, $\mathit{Class}_{uai}$ takes a long time to start producing better results than $\mathit{Class}$. Nevertheless, UAInets are still much more robust than the underlying networks to different types of anomalies, producing good results in all datasets, even when the underlying network produces weak results on the dataset.

\begin{figure*}[htp]
  \centering
    \begin{subfigure}[t]{0.28\textwidth}
        \centering
        \includegraphics[width=\textwidth]{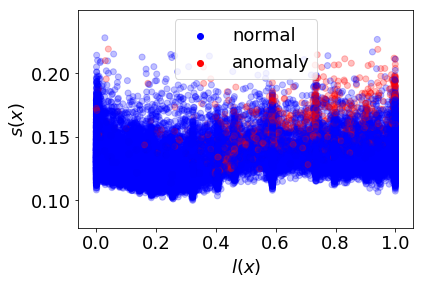}
        \caption{MNIST\textsubscript{0-2} ($b=0$)} \label{fig:hs_vs_loss_0}
    \end{subfigure}%
    ~
    \begin{subfigure}[t]{0.28\textwidth}
        \centering
        \includegraphics[width=\textwidth]{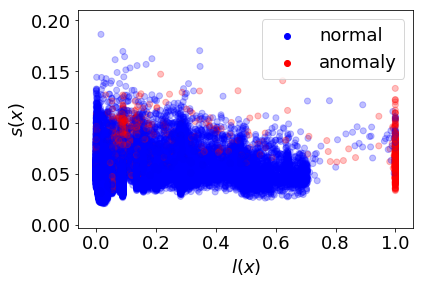}
        \caption{MNIST\textsubscript{0-2} ($b=250$)} \label{fig:hs_vs_loss_2}
    \end{subfigure}%
    ~ 
    \begin{subfigure}[t]{0.28\textwidth}
        \centering
        \includegraphics[width=\textwidth]{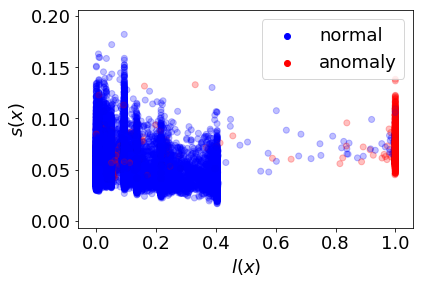}
        \caption{MNIST\textsubscript{0-2} ($b=2000$)} \label{fig:hs_vs_loss_6}
    \end{subfigure}
    ~
    \begin{subfigure}[t]{0.28\textwidth}
        \centering
        \includegraphics[width=\textwidth]{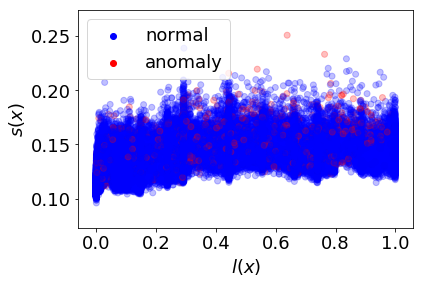}
        \caption{MNIST\textsubscript{hard} ($b=0$)} \label{fig:hs_vs_loss_0:hard}
    \end{subfigure}%
    ~ 
    \begin{subfigure}[t]{0.28\textwidth}
        \centering
        \includegraphics[width=\textwidth]{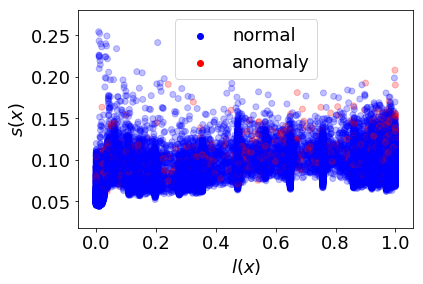}
        \caption{MNIST\textsubscript{hard} ($b=250$)} \label{fig:hs_vs_loss_2:hard}
    \end{subfigure}%
    ~ 
    \begin{subfigure}[t]{0.28\textwidth}
        \centering
        \includegraphics[width=\textwidth]{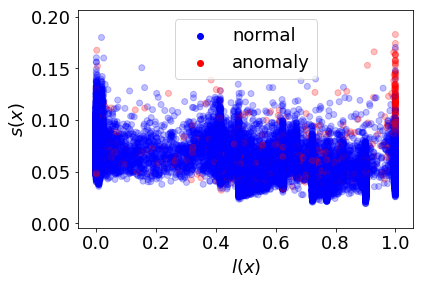}
        \caption{MNIST\textsubscript{hard} ($b=2000$)} \label{fig:hs_vs_loss_6:hard}
    \end{subfigure}
  \caption{(Color online) Underlying latent representations ($l_{DAE}$) vs anomaly score ($s_{DAE}$) for $\mathit{DAE}_{uai}$ network as training progresses.}
  \label{fig:DAE_evolution}
\end{figure*}

\vspace{0.05in}
\noindent\textbf{Learned Representations and Anomaly Scores:} We further analyze UAInets by studying the evolution of hidden representations and anomaly scores through the training process. More specifically, we show visualizations of the learned representations and anomaly scores, presenting their evolution as more labels are fed into the network through the active feedback. With this purpose, we retrain our models on both MNIST\textsubscript{0-2} and MNIST\textsubscript{hard}, with a hidden size of $[256, 64, 1]$, so that their latent representations are one dimensional ($l(x) \in R^1$), and plot these representations versus the anomaly scores of the base network (either $\mathit{DAE}$ or $\mathit{Class}$) for different budgets ($b$).

Figure \ref{fig:DAE_evolution} shows the evolution of $\mathit{DAE}_{uai}$ in terms of $l_{\mathit{DAE}}(x)$ and $s_{\mathit{DAE}}(x)$. We can see from Figures \ref{fig:DAE_evolution}(a) and \ref{fig:DAE_evolution}(d) that initially  anomalies and normal data instances are not separable in this space. Nevertheless, with only a few labeled instances ($b=250$) the anomaly space becomes much easier to separate, while for $b=2000$ the anomaly space is almost perfectly linearly separable.

\begin{figure*}
  \centering
    \begin{subfigure}[t]{0.28\textwidth}
        \centering
        \includegraphics[width=\textwidth]{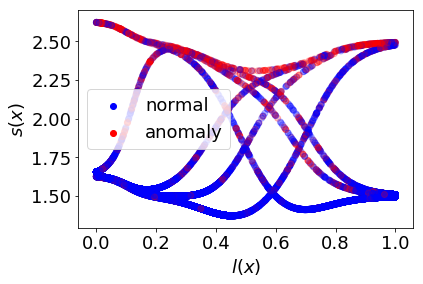}
        \caption{MNIST\textsubscript{0-2} ($b=0$)} \label{fig:hs_vs_loss_0_class}
    \end{subfigure}%
    ~ 
    \begin{subfigure}[t]{0.28\textwidth}
        \centering
        \includegraphics[width=\textwidth]{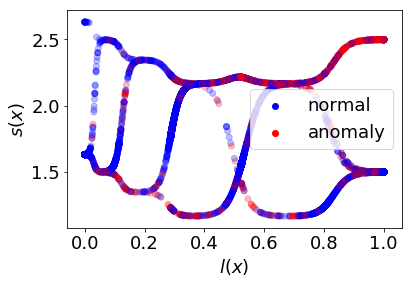}
        \caption{MNIST\textsubscript{0-2} ($b=250$)} \label{fig:hs_vs_loss_2_class}
    \end{subfigure}%
    ~ 
    \begin{subfigure}[t]{0.28\textwidth}
        \centering
        \includegraphics[width=\textwidth]{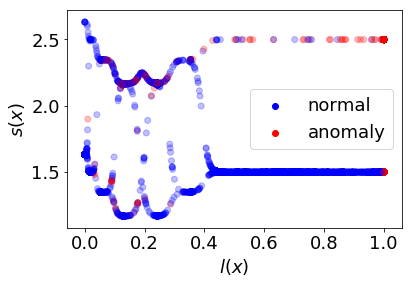}
        \caption{MNIST\textsubscript{0-2} ($b=2000$)} \label{fig:hs_vs_loss_5_class}
    \end{subfigure}
    ~
    \begin{subfigure}[t]{0.28\textwidth}
        \centering
        \includegraphics[width=\textwidth]{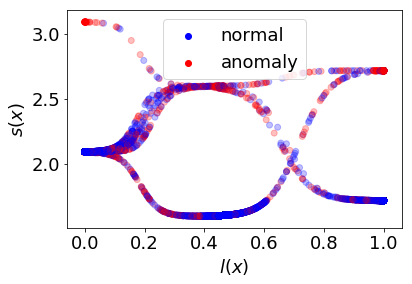}
        \caption{MNIST\textsubscript{hard} ($b=0$)} \label{fig:hs_vs_loss_0_class:hard}
    \end{subfigure}%
    ~ 
    \begin{subfigure}[t]{0.28\textwidth}
        \centering
        \includegraphics[width=\textwidth]{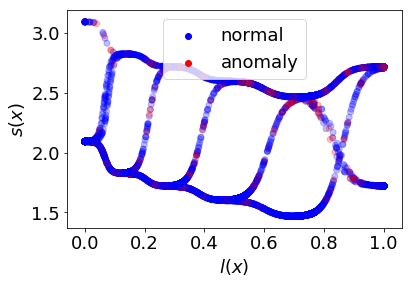}
        \caption{MNIST\textsubscript{hard} ($b=250$)} \label{fig:hs_vs_loss_2_class:hard}
    \end{subfigure}%
    ~ 
    \begin{subfigure}[t]{0.28\textwidth}
        \centering
        \includegraphics[width=\textwidth]{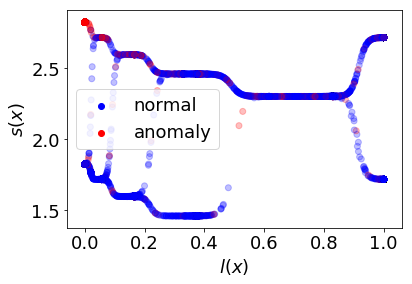}
        \caption{MNIST\textsubscript{hard} ($b=2000$)} \label{fig:hs_vs_loss_6_class:hard}
    \end{subfigure}
  \caption{(Color online) Underlying latent representations ($l_{class}$) vs anomaly score ($s_{class}$) for $\mathit{Class}_{uai}$ network as training progresses.}
  \label{fig:class_evolution}
\end{figure*}

Figure \ref{fig:class_evolution} shows the same evolution for $\mathit{Class}_{uai}$. We can also see the same patterns, as initially anomalies and normal data points are not separable, but with a few labeled instances anomalies become much more identifiable. The main conclusion taken from these visualizations is how the gradient flow through $l$ is important, since it helps the network to better separate data in these anomaly spaces, allowing good anomaly detection performance even when the underlying networks are not good at identifying a specific type of anomaly.

\subsection{Real Data} \label{sec:real_data}

\begin{table*}
	\centering
    \caption{Results on Real Datasets showing average F1 scores of five independent runs.} \label{table:real_results}
\setlength\tabcolsep{5pt}
	\begin{tabular}{lccccccccccc}
	Train Set & KDDCUP & Arrhythmia & Thyroid & KDDCUP-Rev & Yeast & Abalone & CTG & Credit & Covtype & MMG & Shuttle\\
        \hline\\
	\# Instances & $494{,}021$ & $3{,}772$ & 452 & $121{,}597$ & $1{,}191$ & $1,920$ & $1,700$ & $284,807$ & $286,048$ & $11,183$ & $12,345$\\
	\# Features & 120 & 6 & 274 & 120 & 8 & 9 & 22 & 30 & 54 & 6 & 9\\
	\% anomalies & 20\% & $2.5$\% &  15\% & 20\% & 4.6\% & 1.5\% & 2.6\% & 0.17\% & 0.9\% & 2.3\% & 7.0\%\\
        \hline\\
	DAGMM (clean) & \textbf{0.94} & \textbf{0.50} & 0.44 & \textbf{0.94} & 0.11 & 0.16 & 0.27 & 0.34 & 0.18 & 0.07 & 0.48\\
	DAGMM (dirty) & 0.43 & 0.46 & 0.46 & 0.31 & 0.02 & 0.05 & 0.18 & 0.31 & 0.01 & 0.00 & 0.48\\
	LODA-AAD & 0.88 & 0.45 & 0.51 & 0.83 & 0.31 & 0.54 & 0.52 & 0.57 & \textbf{0.97} & 0.42 & \textbf{0.97}\\
	Tree-AAD & 0.89 & 0.29 & \textbf{0.86} & 0.50 & 0.32 &0.53 & \textbf{0.69} & \textbf{0.76} & 0.94 & 0.59 & 0.92\\
	DAE & 0.39 & 0.35 & 0.09 & 0.16 & 0.23 & 0.08 & 0.13 & 0.36 & 0.15 & 0.27 & 0.17\\
	\daeuai{} & \textbf{0.94} & \textbf{0.47} & 0.57 & \textbf{0.91} & \textbf{0.33} & \textbf{0.55} & \textbf{0.66} & 0.64 & 0.86 & \textbf{0.60} & 0.93\\
	\end{tabular}
\end{table*}

We also evaluated the performance of our models on publicly available anomaly detection benchmarks with real anomalies. These datasets were previously considered in \citep{dheeru2017uci}. We compare our model with DAE \citep{vincent2008extracting}, DAGMM \citep{zong2018deep}, LODA-AAD \citep{das2016incorporating}, and Tree-AAD \citep{das2017incorporatingtree}.

Table \ref{table:real_results} presents results for these datasets. In these experiments, DAGMM (clean) was trained on a semi-supervised anomaly detection setting. DAGMM (dirty) and DAE were trained in an unsupervised setting. LODA-AAD, Tree-AAD and $\mathit{DAE}_{uai}$ were trained in an active anomaly detection setting. DAE performs poorly in all the datasets. Nevertheless, even using a simple architecture as its underlying model, $\mathit{DAE}_{uai}$ produces the best performance (or close to the best) on all datasets, even when the baselines were trained in completely clean training sets. $\mathit{DAE}_{uai}$ also usually presents better results than LODA-AAD and Tree-AAD, which were also trained in an active setting.

One possible criticism would be that our results become more relevant the fewer the proportion of anomalous instances, which seems self-defeating. But we see that the largest difference from the active models to the other considered competitors was in Covtype, which has less than 1\% anomalies out of $286{,}048$ instances. When working with large datasets ($>$1M instances), even if only 0.1\% of the dataset is contaminated there is still the chance to benefit from this feedback to improve performance. The active models are also more robust than the others, DAGMM used different hyperparameters for each experiment, while $\mathit{DAE}_{uai}$ and AAD use the same for all (except for k which was reduced from 10 to 3 for the datasets with less than 100 anomalies).

\section{Related Work}

\noindent\textbf{Anomaly Detection:} Although many algorithms have been recently proposed, classical methods for outlier detection like LOF \cite{breunig2000lof} and OC-SVM \citep{scholkopf2001estimating}, are still used in many application scenarios. Recent works have focused on statistical properties of ``normal'' data to identify anomalies, such as \citep{maurus2017let}, which uses Benford's Law to identify anomalies in social networks, and \citep{siffer2017anomaly}, which uses Extreme Value Theory to detect anomalies. Recently, energy based models \citep{zhai2016deep} and GANs \citep{schlegl2017unsupervised} have been successfully used to detect anomalies, but autoencoders are still more popular in this field.
The DAGMM algorithm was proposed in \citep{zong2018deep}, where they train a deep autoencoder and use its latent representations, together with its reconstruction error, as input to a second network, which they use to predict the membership of each data instance to a mixture of gaussian models, training the whole model end-to-end in an semi-supervised manner for novelty detection.

\vspace{0.05in}
\noindent\textbf{Active Anomaly Detection:} Over the years interesting works have been developed in this topic. In \citep{pelleg2005active}, the authors solve the rare-category detection problem by proposing an active learning strategy to datasets with extremely skewed distributions. In~\citet{abe2006outlier}, the authors reduce outlier detection to classification using artificially generated examples that play the role of potential outliers and then applies a selective sampling mechanism based on active learning to the reduced classification problem. In \citep{gornitz2013toward}, the authors propose a Semi-Supervised Anomaly Detection (SSAD) method based on Support Vector Data Description (SVDD) \citep{tax2004support}. In~\citet{veeramachaneni2016ai}, the authors propose an active approach that combines unsupervised and supervised learning to select instances to be labeled by experts. 
In~\citet{sharma2016active}, they use an active learning approach to identify significant anomalies in aviation. They require explanations on expert annotators' choices, which they use to iteratively create new features with which they improve their model.
The most similar works to ours in this setting are (i) \citep{das2016incorporating}, which proposes an algorithm that can be employed on top of any ensemble methods based on random projections, and (ii) \citep{das2017incorporatingtree}, which expands Isolation Forests to work in an active setting.
\section{Conclusions and Future Work}

In this work we proposed an Unsupervised to Active Inference layer (or simply UAI layer) that can be applied on top of any deep learning architecture designed for unsupervised anomaly detection. We showed that, even on top of very simple architectures like autoencoders and multi-layer perceptrons, our models achieve similar or better results than state-of-the-art representatives. To the best of our knowledge, this is the first work which applies deep learning to active anomaly detection. We used the strongest points of unsupervised deep learning solutions (learned representations and anomaly scores) to transform them into active learning models, presenting an end-to-end architecture which learns representations by leveraging both the full dataset and the already labeled instances.

Important future directions for this work are: (i) using the UAI layers confidence in its output to dynamically choose between either directly using its scores, or using the underlying unsupervised model's anomaly score to choose which instances to audit next; (ii) testing new architectures for UAI layers, in this work we restricted all our analysis to simple logistic regression; (iii) analyzing the robustness of UAINets to mistakes being made by the labeling experts; (iv) and making this model more interpretable, so that auditors could focus on a few ``important'' features when labeling anomalous instances, which could increase labeling speed and make their work easier.
 
\bibliographystyle{IEEEtran}
\bibliography{anomalies}

\end{document}